\documentclass[runningheads]{llncs}
\usepackage[T1]{fontenc}
\usepackage{graphicx,verbatim}
\usepackage{multicol,multirow}
\usepackage{hyperref}
\usepackage{amsmath}
\usepackage{amssymb}
\usepackage{makecell}
\usepackage{algorithm}
\usepackage{algpseudocode}

\begin{document}
\title{Unsupervised Out-of-Distribution Detection in Medical Imaging Using Multi-Exit Class Activation Maps and Feature Masking}

\author{Yu-Jen Chen, Xueyang Li, Yiyu Shi, and Tsung-Yi Ho}  
\authorrunning{Chen et al.}
\institute{National Tsing Hua University, Taiwan \\
    \email{chenjuzen@gmail.com}
\and University of Notre Dame, Notre Dame, IN, USA\\
    \email{\{xli34, yshi4\}@nd.edu}
\and The Chinese University of Hong Kong, Hong Kong\\
    \email{tyho@cse.cuhk.edu.hk}}
\titlerunning{Unsupervised Medical OOD using MECAM}
    
\maketitle              
\begin{abstract}
Out-of-distribution (OOD) detection is essential for ensuring the reliability of deep learning models in medical imaging applications. This work is motivated by the observation that class activation maps (CAMs) for in-distribution (ID) data typically emphasize regions that are highly relevant to the model’s predictions, whereas OOD data often lacks such focused activations. By masking input images with inverted CAMs, the feature representations of ID data undergo more substantial changes compared to those of OOD data, offering a robust criterion for differentiation.
In this paper, we introduce a novel unsupervised OOD detection framework, Multi-Exit Class Activation Map (MECAM), which leverages multi-exit CAMs and feature masking. By utilizing multi-exit networks that combine CAMs from varying resolutions and depths, our method captures both global and local feature representations, thereby enhancing the robustness of OOD detection. We evaluate MECAM on multiple ID datasets, including ISIC19 and PathMNIST, and test its performance against three medical OOD datasets, RSNA Pneumonia, COVID-19, and HeadCT, and one natural image OOD dataset, iSUN. Comprehensive comparisons with state-of-the-art OOD detection methods validate the effectiveness of our approach. Our findings emphasize the potential of multi-exit networks and feature masking for advancing unsupervised OOD detection in medical imaging, paving the way for more reliable and interpretable models in clinical practice. Our code is available at \url{https://github.com/windstormer/MECAM-OOD}.


\end{abstract}
\section{Introduction}
\label{sec:intro}

Out-of-distribution (OOD) detection plays a pivotal role in enhancing the reliability and safety of deep learning models, especially in high-stakes domains such as medical imaging. In real-world clinical scenarios, models often encounter inputs that deviate from the data distribution they were trained on, posing risks to patient safety if not identified correctly. Detecting OOD inputs is essential to ensure that models make reliable predictions and abstain from making predictions on data outside their scope of expertise.

Existing OOD detection methods can be broadly categorized into softmax score-based methods and network-derived scoring functions \cite{graham2023unsupervised,lemar2025typicality,linmans2024diffusion,liu2023unsupervised,mishra2023dual}. Early approaches, such as MSP \cite{hendrycks2016baseline} and ODIN \cite{liang2017enhancing}, rely on softmax confidence scores to detect OOD samples, while energy-based scoring \cite{liu2020energy} refines this approach by using an energy-derived score. However, these methods often produce overconfident predictions and lack spatial awareness, making them less effective for detecting small, localized OOD patterns in medical images.

To address these limitations, network-derived scoring functions have been proposed. For instance, MOOD \cite{lin2021mood} introduces multi-exit networks to compute energy-based OOD scores, while FeatureNorm \cite{yu2023block} and CORES \cite{tang2024cores} leverage intermediate feature responses for OOD detection. However, these methods assume that the network will provide distinct activation strengths and frequencies for ID and OOD data, which may not always hold true in medical imaging.

Class activation maps (CAMs) offer a potential solution by providing spatially resolved insights into the regions contributing to model predictions. However, single-exit networks, which generate CAMs only from the final layer, often produce low-resolution activation maps due to repeated downscaling. Extracting CAMs from intermediate layers is beneficial because it captures both local and global features that are essential for robust OOD detection.

\begin{figure}[h]
\centering
\includegraphics[width=0.95\linewidth]{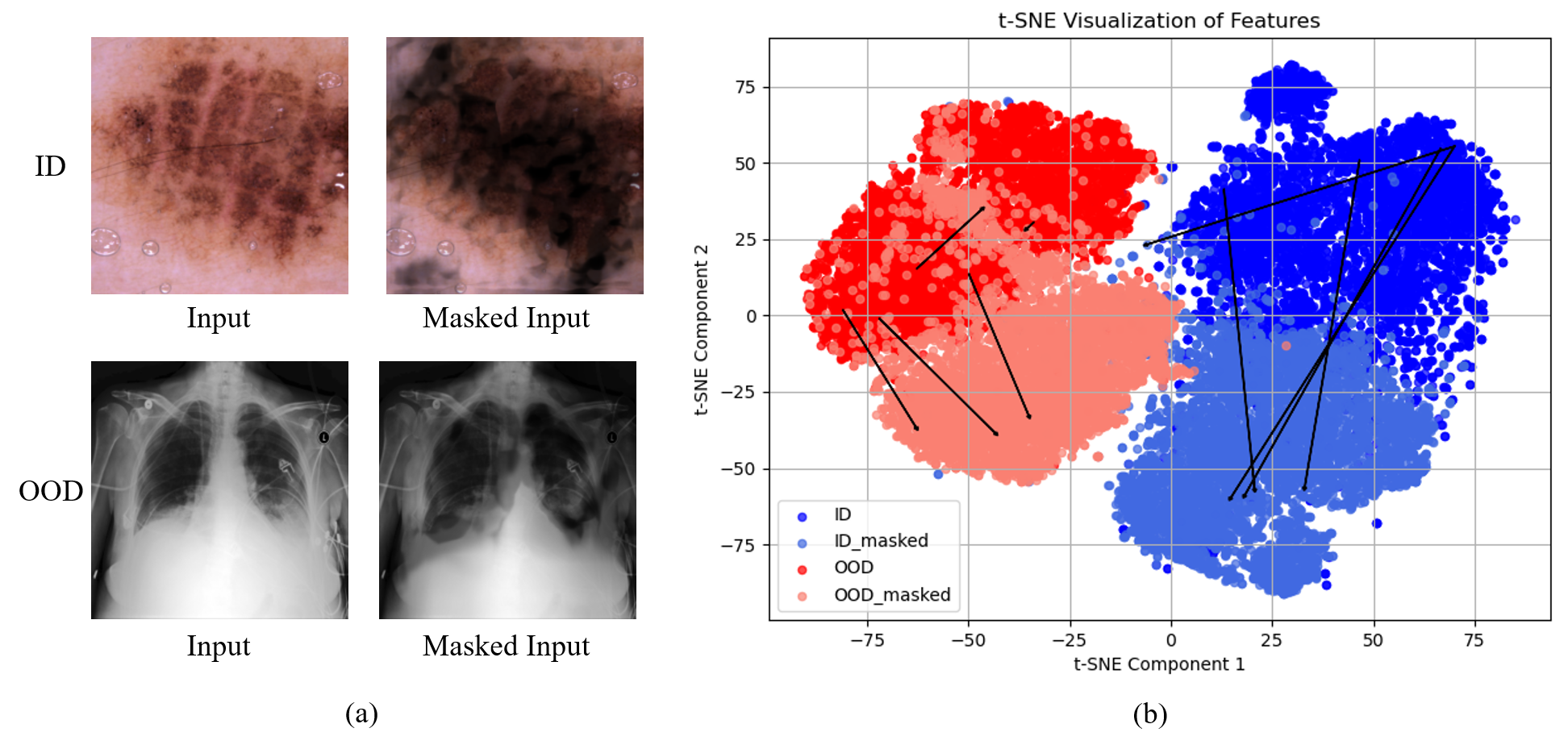}
\caption{Illustration showing that masking the image with the CAM produces significant changes in feature representations. (a) Example of ID and OOD images with their corresponding masked images. (b) Visualization of the features of images and masked images for both ID and OOD data. ID: ISIC dataset, OOD: RSNA Pneumonia dataset.}
\label{fig:overall}
\end{figure}

Motivated by these challenges, we propose Multi-Exit Class Activation Map (MECAM), a novel unsupervised OOD detection framework that leverages multi-exit networks and feature masking. By generating CAMs at different network depths, MECAM effectively captures hierarchical feature representations, combining information from multiple resolutions. Additionally, our feature masking strategy suppresses ID regions using inverted CAMs, amplifying the differences between ID and OOD data. As illustrated in Fig. \ref{fig:overall}, ID images exhibit larger feature changes when masked compared to OOD images, further reinforcing our approach.

We evaluate MECAM on two diverse ID datasets, ISIC19 and PathMNIST, and compare its performance against four OOD datasets, including three medical datasets (RSNA Pneumonia, COVID-19, and HeadCT) and one natural image dataset (iSUN). Extensive experiments demonstrate that MECAM consistently outperforms state-of-the-art OOD detection methods, highlighting its robustness in both medical and natural imaging contexts.

The contributions of this paper are summarized as follows:
\begin{itemize}
    \item We introduce MECAM, a novel framework for unsupervised OOD detection that integrates multi-exit CAMs and feature masking.
    \item To the best of the authors' knowledge, this is the first OOD detection framework that leverages CAMs for unsupervised OOD detection.
    \item We demonstrate that combining CAMs from different depths and resolutions enhances the model’s ability to capture both global and local feature representations.
    \item We conduct comprehensive experiments across multiple ID and OOD datasets, showcasing the effectiveness and generalizability of our approach in diverse imaging scenarios.
\end{itemize}


\section{MECAM-OODD}
\label{sec:methodology}
\begin{figure}[h]
\centering
\includegraphics[width=0.8\linewidth]{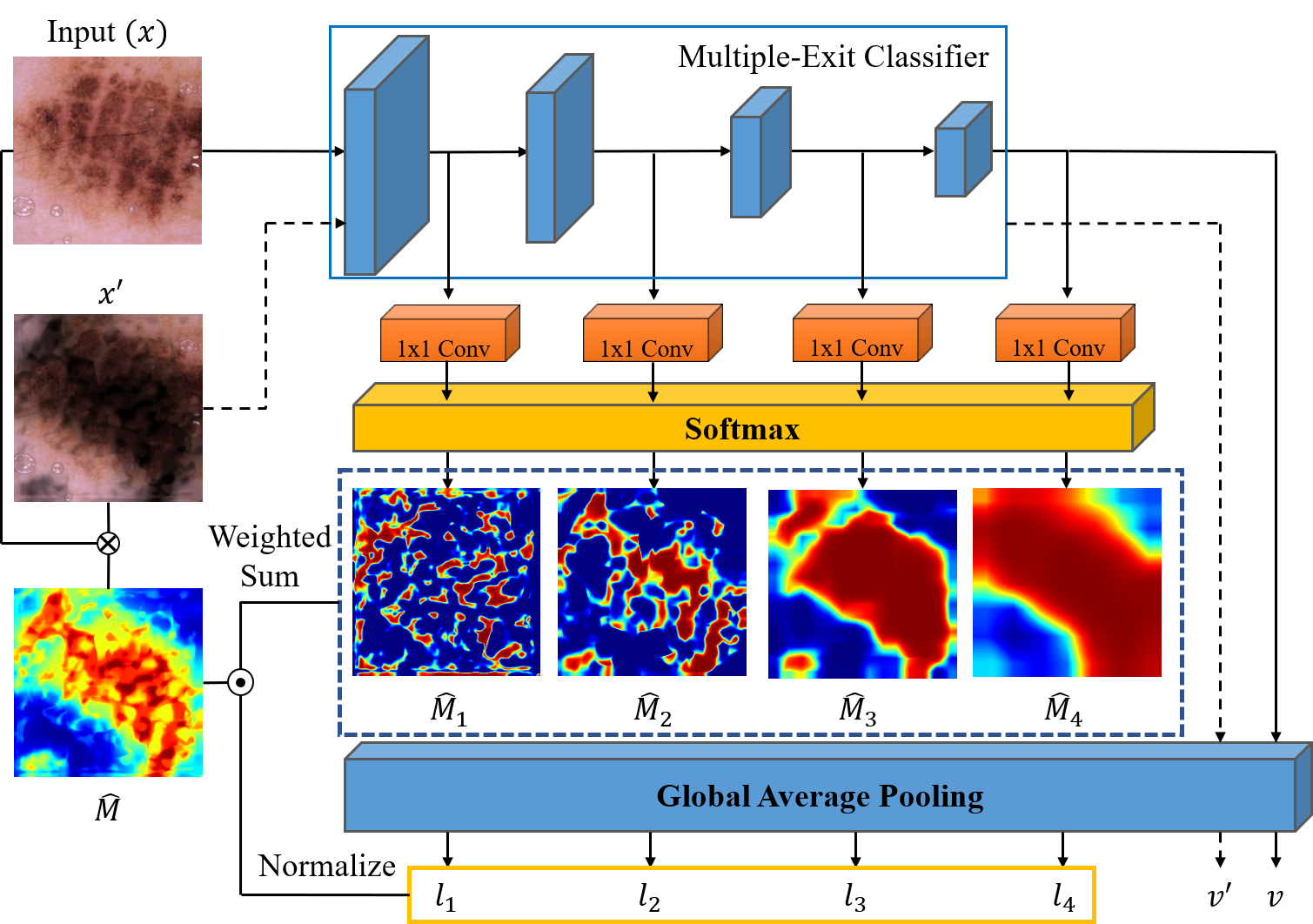}
\caption{An illustration of the proposed MECAM framework for OOD detection.}
\label{fig:framework}
\end{figure}

To address the limitations of existing out-of-distribution (OOD) detection methods in medical imaging, we propose MECAM, a multi-exit class activation map (CAM)-based approach. This method is motivated by the observation that in-distribution (ID) data exhibits significant changes in feature representations when masked, whereas OOD data remains relatively unaffected. MECAM leverages spatial information from multiple network depths by extracting CAMs from various layers, aggregating them using a weighted scheme, and applying feature masking to enhance OOD detection. An overview of the framework is provided in Fig. \ref{fig:framework}.

As shown in Fig. \ref{fig:framework}, the model is built on a multi-exit network that generates CAMs at various intermediate layers. Each exit corresponds to a convolutional layer followed by a classification head, allowing for the extraction of CAMs at different levels of abstraction. Following \cite{chen2023ame}, we train the multi-exit classifier using a multi-exit cross-entropy loss, which combines the cross-entropy losses from each exit using a weighted sum.

Given an input image $x$, we obtain the activation map $M_e$ and output logit $l_e$ at exit $e$ by inferring the model $f(x)$. The predicted class $P$ is defined as the class with the largest output logit at the final exit.

In the initial step of CAM extraction, the CAM $\hat{M}$ at exit $e$ for class $c$ is computed by applying the softmax function along the class dimension. We then select the CAM corresponding to the predicted class $P$ at each exit, resulting in $\hat{M}_{e,P}(x)$, which is abbreviated as $\hat{M}_e(x)$.

To effectively combine the CAMs from different exits, we compute a weight for each exit based on the normalized classification logit across all exits. This weighting scheme ensures that exits with higher logits (i.e., higher predicted confidence), which typically capture more relevant features, contribute more significantly to the final CAM $\hat{M}(x)$.

Using the final CAM, we generate a masked image $x'$ by suppressing the regions relevant to in-distribution data:
\[
x' = x \otimes \left(1 - \hat{M}(x)\right).
\]

To obtain the image embedding, we use the output from the layer immediately preceding the final convolutional layer. Thus, the embedding of the original image, $v = f(x)$, and the embedding of the masked image, $v' = f(x')$, are obtained by inferring the model $f$. The OOD score is calculated using the mean squared error (MSE) based on the feature shift induced by masking:
\[
\text{Score}_{OOD} = \frac{1}{d} \sum_{i=1}^{d} \left(v_i - v'_i\right)^2.
\]

A larger feature shift indicates that the original features relied heavily on the masked regions, suggesting that the input is in-distribution. Conversely, a smaller shift indicates OOD characteristics.

In conclusion, an input image $x$ is classified as ID or OOD based on the OOD score using a threshold $\tau$:
\begin{equation}
g_{\tau}(x; f) =
\begin{cases}
\text{ID} & \text{if } \text{Score}_{OOD} \geq \tau, \\
\text{OOD} & \text{otherwise,}
\end{cases}
\end{equation}
where the threshold $\tau$ is typically chosen such that 95\% of ID data is correctly classified as ID (i.e., a true positive rate of 95\%), and $f$ denotes the multi-exit classifier.

\section{Experiments}
\label{sec:experiments}
In this paper, we use ISIC \cite{codella2018skin,combalia2019bcn20000,tschandl2018ham10000} and PathMNIST \cite{kather2019predicting} as our in-distribution (ID) datasets. ISIC is a large-scale dermoscopic image dataset for skin lesion classification, containing approximately 25,000 training images and 6,191 test images. PathMNIST, derived from the MedMNIST \cite{yang2023medmnist} collection, consists of histopathology images categorized into nine tissue types, with 89,996 training images and 7,180 test images.

For out-of-distribution (OOD) detection, we create a mixed test set by combining the test set of an ID dataset with that of an OOD dataset, in order to determine whether each instance belongs to the ID or OOD category. The OOD datasets include three medical datasets—RSNA Pneumonia \cite{shih2019augmenting} (5,337 images), COVID-19 \cite{chowdhury2020can,rahman2021exploring} (13,808 images), and HeadCT \cite{Kitamura2018HeadCT} (200 images)—as well as iSUN \cite{xu2015turkergaze} (8,181 images), a natural image dataset. These OOD datasets contain images from domains distinct from those of the in-distribution datasets.

Following previous works \cite{tang2024cores,yu2023block}, we evaluate OOD detection performance using the Area Under the Receiver Operating Characteristic Curve (AUC) and the False Positive Rate at 95\% True Positive Rate (FPR95), both reported as percentages.

All models were trained using PyTorch on two RTX 3080 Ti 12 GB GPUs. During training, input images are resized to 224 $\times$ 224 with a batch size of 128. The model is optimized using the SGD optimizer with an initial learning rate of 0.01 that decays to $1\times10^{-4}$. For the ISIC and PathMNIST datasets, the model is trained for 200 epochs and 20 epochs, respectively. To prevent overfitting, a weight decay of $1\times10^{-4}$ is applied, and we follow the data augmentation protocols described in \cite{chiu2024achieve}.

\section{Results}
\label{sec:results}
\subsection{Comparison with State-of-the-Art}

\begin{table}[h]
\centering
\scriptsize
\caption{Comparison of OOD detection performance among different methods under a small-scale setting. }
\setlength{\tabcolsep}{1pt}
\label{Table:sota}
\begin{tabular}{c|c|c|cccccccc}
\hline
\multirow{3}{*}{ID}         & \multirow{3}{*}{Model} & \multirow{3}{*}{Method} & \multicolumn{8}{c}{OOD}                                                                                                                                   \\ \cline{4-11} 
                            &                        &                         & \multicolumn{2}{c}{RSNA}             & \multicolumn{2}{c}{COVID-19}         & \multicolumn{2}{c}{HeadCT}           & \multicolumn{2}{c}{iSUN}             \\ \cline{4-11} 
                            &                        &                         & FPR95 $\downarrow$ & AUC $\uparrow$  & FPR95 $\downarrow$ & AUC $\uparrow$  & FPR95 $\downarrow$ & AUC $\uparrow$  & FPR95 $\downarrow$ & AUC $\uparrow$  \\ \hline
\multirow{16}{*}{\rotatebox{90}{ISIC19}}    & \multirow{8}{*}{\rotatebox{90}{ResNet-18}} & MSP \cite{hendrycks2016baseline}                     & 79.05             & 74.63          & 86.35             & 68.22          & 60.68             & 92.50          & 59.30             & 83.60          \\
                            &                        & ODIN \cite{liang2017enhancing}                    & 90.58             & 66.27          & 96.85             & 57.42          & 63.45             & 90.00          & 37.26             & 89.60          \\
                            &                        & Energy \cite{liu2020energy}                  & 50.83             & 89.87          & 61.65             & 87.38          & 73.27             & 85.00          & 49.14             & 88.88          \\
                            &                        & MOOD \cite{lin2021mood}                    & 99.91             & 72.83          & 100.00            & 58.90          & 92.50             & 72.77          & 99.99             & 57.60          \\
                            &                        & DICE \cite{sun2022dice}                    & 82.21             & 80.83          & 89.96             & 70.97          & 39.00             & 93.48          & 46.57             & 84.86          \\
                            &                        & FeatureNorm\cite{yu2023block}             & 55.91             & 90.46          & 60.81             & 87.45          & 24.50             & 95.89          & 57.14             & 80.10          \\
                            &                        & CORES \cite{tang2024cores}                   & 38.50             & 92.78          & 26.87             & 94.58          & 59.00             & 90.92          & 72.67             & 76.83          \\
                            &                        & Ours                    & \textbf{00.62}    & \textbf{99.62} & \textbf{13.71}    & \textbf{97.66} & \textbf{06.00}    & \textbf{98.78} & \textbf{17.11}    & \textbf{95.89} \\ \cline{2-11} 
                            & \multirow{8}{*}{\rotatebox{90}{ResNet-50}} & MSP \cite{hendrycks2016baseline}                     & 99.44             & 54.66          & 97.98             & 56.30          & 81.00             & 76.32          & 65.35             & 82.37          \\
                            &                        & ODIN \cite{liang2017enhancing}                    & 99.19             & 54.68          & 98.67             & 46.72          & 70.50             & 76.21          & 46.57             & 86.96          \\
                            &                        & Energy \cite{liu2020energy}                  & 99.36             & 60.10          & 97.48             & 60.73          & 72.50             & 82.30          & 46.54             & 87.18          \\
                            &                        & MOOD \cite{lin2021mood}                    & 98.80             & 74.96          & 99.92             & 58.50          & 97.50             & 65.17          & 99.62             & 55.12          \\
                            &                        & DICE \cite{sun2022dice}                    & 99.63             & 67.06          & 99.31             & 62.35          & 55.00             & 92.37          & 81.55             & 75.59          \\
                            &                        & FeatureNorm\cite{yu2023block}             & 100.00            & 23.12          & 99.78             & 30.22          & 100.00            & 38.64          & 99.51             & 26.88          \\
                            &                        & CORES \cite{tang2024cores}                   & 98.78             & 67.44          & 99.58             & 67.20          & 94.50             & 72.84          & 95.36             & 37.19          \\
                            &                        & Ours                    & \textbf{02.90}    & \textbf{99.20} & \textbf{13.41}    & \textbf{97.60} & \textbf{24.50}    & \textbf{95.13} & \textbf{31.99}    & \textbf{93.06} \\ \hline
\multirow{16}{*}{\rotatebox{90}{PathMNIST}} & \multirow{8}{*}{\rotatebox{90}{ResNet-18}} & MSP \cite{hendrycks2016baseline}                     & 57.80             & 93.10          & 51.86             & 93.77          & 47.00             & 90.44          & 74.07             & 84.50          \\
                            &                        & ODIN \cite{liang2017enhancing}                    & 67.28             & 86.20          & 62.01             & 88.10          & 80.50             & 56.77          & 46.25             & 89.21          \\
                            &                        & Energy \cite{liu2020energy}                  & 26.59             & 95.66          & 28.08             & 95.56          & 48.50             & 88.07          & 51.84             & 88.45          \\
                            &                        & MOOD \cite{lin2021mood}                    & 15.78             & 97.21          & 66.06             & 93.73          & \textbf{03.00}    & \textbf{99.34} & 39.24             & 92.13          \\
                            &                        & DICE \cite{sun2022dice}                    & 100.00            & 39.54          & 100.00            & 42.64          & 98.50             & 57.86          & 99.98             & 56.10          \\
                            &                        & FeatureNorm\cite{yu2023block}             & 99.91             & 84.17          & 99.67             & 85.05          & 100.00            & 82.69          & 99.74             & 79.87          \\
                            &                        & CORES \cite{tang2024cores}                   & 100.00            & 39.48          & 99.82             & 39.81          & 100.00            & 51.74          & 99.83             & 66.81          \\
                            &                        & Ours                    & \textbf{06.82}    & \textbf{98.38} & \textbf{13.60}    & \textbf{97.54} & 06.00             & 98.21          & \textbf{01.04}    & \textbf{99.17} \\ \cline{2-11} 
                            & \multirow{8}{*}{\rotatebox{90}{ResNet-50}} & MSP \cite{hendrycks2016baseline}                     & 15.72             & 97.52          & 18.03             & 96.92          & 33.00             & 93.66          & 95.21             & 56.04          \\
                            &                        & ODIN \cite{liang2017enhancing}                    & 25.69             & 94.43          & 41.66             & 86.10          & 71.00             & 74.03          & 89.18             & 60.45          \\
                            &                        & Energy \cite{liu2020energy}                  & 06.33             & 98.55          & 09.98             & 97.68          & 32.00             & 91.39          & 90.89             & 54.93          \\
                            &                        & MOOD \cite{lin2021mood}                    & 48.17             & 94.16          & 82.39             & 89.75          & 21.50             & 96.28          & 72.96             & 85.02          \\
                            &                        & DICE \cite{sun2022dice}                    & 100.00            & 49.39          & 99.99             & 54.94          & 93.50             & 57.88          & 99.98             & 65.80          \\
                            &                        & FeatureNorm\cite{yu2023block}             & 99.98             & 20.48          & 99.77             & 26.92          & 96.00             & 77.28          & 97.36             & 56.56          \\
                            &                        & CORES \cite{tang2024cores}                   & 100.00            & 48.82          & 99.90             & 56.96          & 100.00            & 37.12          & 99.94             & 86.23          \\
                            &                        & Ours                    & \textbf{00.30}    & \textbf{99.69} & \textbf{02.80}    & \textbf{99.06} & \textbf{03.50}    & \textbf{98.88} & \textbf{04.29}    & \textbf{98.45} \\ \hline
\end{tabular}
\end{table}

Our experiments compare the proposed MECAM with baseline OOD detection methods such as MSP \cite{hendrycks2016baseline}, ODIN \cite{liang2017enhancing}, Energy \cite{liu2020energy}, DICE \cite{sun2022dice}, FeatureNorm \cite{yu2023block}, and CORES \cite{tang2024cores}. Additionally, we compare MECAM with MOOD \cite{lin2021mood}, which also employs a multi-exit network.

Table \ref{Table:sota} presents the results of various OOD detection methods evaluated on the ISIC19 and PathMNIST datasets, using RSNA, COVID-19, HeadCT (medical OOD datasets), and iSUN (a natural image OOD dataset). To ensure a fair comparison, we conduct experiments using both ResNet-18 and ResNet-50 as backbone architectures, thereby assessing the scalability and effectiveness of MECAM across different model capacities.

For ISIC19, MECAM consistently outperforms the baseline approaches across medical OOD datasets. Specifically, MECAM achieves a significant reduction in FPR95, with values on average 6$\times$ and 3$\times$ lower than the baselines for ResNet-18 and ResNet-50, respectively. In addition, MECAM demonstrates an AUC improvement of 32.93\% for ResNet-18 and 54.95\% for ResNet-50. 

Similarly, for PathMNIST, MECAM surpasses state-of-the-art methods across medical OOD datasets, achieving an average FPR95 reduction of 9$\times$ for ResNet-18 and 24$\times$ for ResNet-50. Moreover, MECAM improves AUC by 92.14\% and 244.51\% for ResNet-18 and ResNet-50, respectively. These results underscore the effectiveness of MECAM in reducing false positives and enhancing OOD discrimination in medical imaging scenarios.

MECAM also achieves strong performance when distinguishing between medical ID datasets (ISIC19, PathMNIST) and the natural image OOD dataset (iSUN). On ISIC19, MECAM reduces FPR95 by an average of 4.2$\times$ for ResNet-18 and 2.5$\times$ for ResNet-50, while improving AUC by 15.64\% and 22.91\%, respectively. For PathMNIST, MECAM attains an FPR95 reduction of 8.7$\times$ for ResNet-18 and 19.8$\times$ for ResNet-50, along with AUC improvements of 37.85\% and 48.26\%, respectively. These results suggest that MECAM is highly effective at distinguishing medical data from out-of-domain natural images, further demonstrating its generalizability to diverse OOD scenarios.

The consistent performance improvement across different datasets and OOD scenarios underscores MECAM's robustness and adaptability in medical imaging applications.

\subsection{Comparison with Different CAM Approaches}
\begin{table}[h]
\centering
\scriptsize
\caption{Comparison of OOD detection performance among different methods employing various CAM-based approaches. Res18 and Res50 denote ResNet-18 and ResNet-50, respectively.}
\setlength{\tabcolsep}{1pt}
\label{Table:ablation_CAM}
\begin{tabular}{c|c|c|cccccccc}
\hline
\multirow{3}{*}{ID} & \multirow{3}{*}{Model} & \multirow{3}{*}{Method} & \multicolumn{8}{c}{OOD} \\ \cline{4-11} 
 &  &  & \multicolumn{2}{c}{RSNA} & \multicolumn{2}{c}{COVID-19} & \multicolumn{2}{c}{HeadCT} & \multicolumn{2}{c}{iSUN} \\ \cline{4-11} 
 &  &  & FPR95 $\downarrow$ & AUC $\uparrow$ & FPR95 $\downarrow$ & AUC $\uparrow$ & FPR95 $\downarrow$ & AUC $\uparrow$ & FPR95 $\downarrow$ & AUC $\uparrow$ \\ \hline
\multirow{6}{*}{\rotatebox{90}{ISIC19}} & \multirow{3}{*}{\rotatebox{90}{Res18}} & Grad-CAM \cite{selvaraju2017grad} & 46.28 & 91.19 & 64.74 & 80.96 & 52.00 & 90.28 & 45.89 & 87.81 \\
 &  & LayerCAM \cite{jiang2021layercam} & 26.83 & 95.80 & 54.48 & 87.51 & 61.50 & 87.18 & 35.08 & 92.16 \\
 &  & Ours & \textbf{00.62} & \textbf{99.62} & \textbf{13.71} & \textbf{97.66} & \textbf{06.00} & \textbf{98.78} & \textbf{17.11} & \textbf{95.89} \\ \cline{2-11} 
 & \multirow{3}{*}{\rotatebox{90}{Res50}} & Grad-CAM \cite{selvaraju2017grad} & 93.97 & 65.92 & 96.37 & 57.16 & 83.50 & 77.47 & 76.75 & 75.38 \\
 &  & LayerCAM \cite{jiang2021layercam} & 57.95 & 91.26 & 75.77 & 81.45 & 50.00 & 90.82 & 50.47 & 85.93 \\
 &  & Ours & \textbf{02.90} & \textbf{99.20} & \textbf{13.41} & \textbf{97.60} & \textbf{24.50} & \textbf{95.13} & \textbf{31.99} & \textbf{93.06} \\ \hline
\multirow{6}{*}{\rotatebox{90}{PathMNIST}} & \multirow{3}{*}{\rotatebox{90}{Res18}} & Grad-CAM \cite{selvaraju2017grad} & 83.79 & 66.17 & 90.72 & 59.88 & 78.00 & 74.87 & 74.09 & 82.72 \\
 &  & LayerCAM \cite{jiang2021layercam} & 97.30 & 74.28 & 97.31 & 72.18 & 91.00 & 63.46 & 92.14 & 60.30 \\
 &  & Ours & \textbf{06.82} & \textbf{98.38} & \textbf{13.60} & \textbf{97.54} & \textbf{06.00} & \textbf{98.21} & \textbf{01.04} & \textbf{99.17} \\ \cline{2-11} 
 & \multirow{3}{*}{\rotatebox{90}{Res50}} & Grad-CAM \cite{selvaraju2017grad} & 84.24 & 74.71 & 83.34 & 73.66 & 82.50 & 70.97 & 43.19 & 92.82 \\ 
 &  & LayerCAM \cite{jiang2021layercam} & 82.89 & 77.27 & 77.01 & 79.69 & 94.00 & 60.38 & 73.21 & 87.89 \\
 &  & Ours & \textbf{00.30} & \textbf{99.69} & \textbf{02.80} & \textbf{99.06} & \textbf{03.50} & \textbf{98.88} & \textbf{04.29} & \textbf{98.45} \\ \hline
\end{tabular}
\end{table}

To further evaluate the effectiveness of MECAM, we compare it with different CAM-based approaches, including Grad-CAM \cite{selvaraju2017grad} and LayerCAM \cite{jiang2021layercam}.

As shown in Table \ref{Table:ablation_CAM}, across all OOD datasets, MECAM consistently outperforms Grad-CAM and LayerCAM, achieving lower FPR95 and higher AUC. On ISIC19, MECAM reduces FPR95 by up to 4$\times$ and improves AUC by 9.96\% and 23.12\% for ResNet-18 and ResNet-50, respectively. Similarly, on PathMNIST, MECAM lowers FPR95 by 12$\times$ and 27$\times$, while improving AUC by 41.53\% and 24.72\% for ResNet-18 and ResNet-50, respectively, demonstrating its effectiveness in enhancing OOD discrimination. Unlike Grad-CAM and LayerCAM, which rely on a single-layer response, MECAM effectively captures hierarchical feature variations, leading to more robust OOD detection.


\subsection{Ablation Study}

\begin{table}[h]
\centering
\caption{Ablation study on the impact of using different exits in the proposed method on OOD detection performance.}
\label{Table:ablation_exit}
\begin{tabular}{c|c|cccc|cc}
\hline
\multirow{2}{*}{ID} & \multirow{2}{*}{Model} & \multicolumn{4}{c|}{Exit} & \multicolumn{2}{c}{OOD RSNA} \\ \cline{3-8} 
 &  & Exit1 & Exit2 & Exit3 & Exit4 & FPR95 $\downarrow$ & AUC $\uparrow$ \\ \hline
\multirow{15}{*}{\rotatebox{90}{ISIC19}} & \multirow{15}{*}{\rotatebox{90}{ResNet-50}} & v &  &  &  & 56.10 & 93.03 \\
 &  &  & v &  &  & 29.17 & 95.38 \\
 &  &  &  & v &  & 07.25 & 98.62 \\
 &  &  &  &  & v & 01.37 & 99.49 \\ \cline{3-8} 
 &  & v & v &  &  & 15.01 & 97.41 \\
 &  & v &  & v &  & 08.77 & 98.12 \\
 &  & v &  &  & v & 08.00 & 98.08 \\
 &  &  & v & v &  & 03.65 & 99.11 \\
 &  &  & v &  & v & 01.24 & 99.39 \\
 &  &  &  & v & v & 00.75 & 99.66 \\ \cline{3-8} 
 &  & v & v & v &  & 01.80 & 99.37 \\
 &  & v & v &  & v & 02.10 & 99.17 \\
 &  & v &  & v & v & 00.58 & 99.55 \\
 &  &  & v & v & v & 00.41 & 99.68 \\ \cline{3-8} 
 &  & v & v & v & v & \textbf{00.30} & \textbf{99.69} \\ \hline
\end{tabular}
\end{table}

Table \ref{Table:ablation_exit} presents the OOD detection performance of MECAM using different exit combinations. The results indicate that relying on individual exits yields suboptimal performance, with earlier exits exhibiting higher FPR95 and lower AUC. The best performance is achieved when all exits are utilized, reducing FPR95 to 0.30\% and achieving an AUC of 99.69\%, which demonstrates the importance of multi-exit feature aggregation for robust OOD detection.

These findings highlight the advantages of leveraging multiple exits in MECAM. By extracting and combining features from different network depths, the model is better able to capture both local and global representations, leading to more accurate and reliable OOD detection.

\section{Conclusion}
\label{sec:conclusion}
In this work, we introduced MECAM, a multi-exit class activation map (CAM)-based approach for out-of-distribution (OOD) detection in medical imaging. By leveraging CAMs extracted from multiple network depths and aggregating them using a confidence-weighted scheme, our method effectively enhances OOD discrimination. Moreover, our feature masking strategy further improves the separation between in-distribution and out-of-distribution data. Extensive experiments on the ISIC19 and PathMNIST datasets across various OOD benchmarks demonstrate that MECAM consistently outperforms state-of-the-art methods. Overall, MECAM provides a more robust and interpretable solution for OOD detection and can be extended to various medical imaging applications where reliable OOD detection is critical.

\bibliographystyle{splncs04}
\bibliography{miccai25,dataset}

\end{document}